# FinRCA-Bench: Benchmarking Evidence Retrieval and Reasoning for Financial AI Systems

Pratik Ghawate

Researcher

## Abstract

Automated financial operations increasingly depend on language models to diagnose why a transaction failed to reconcile, but the evidence that establishes such a diagnosis is rarely contained in a document: it is distributed across invoices, purchase orders, approvals, allocations, payments, ledger entries, and bank activity, and it is connected by transactional relationships rather than by textual similarity. Evaluations that score only the final answer therefore conflate two distinct capabilities - whether a system can find the required evidence and whether it can reason over it. We introduce FinRCA-Bench, a deterministic synthetic benchmark of 2,250 accounts-payable-to-bank reconciliation cases spanning 14 operational tables, comprising 1,500 injected failures across 15 causal categories and 750 legitimate or hard-negative cases. The benchmark separates model-visible operational data from evaluator-private root-cause labels and record-level evidence contracts, so that evidence access is scored independently of answer correctness. We evaluate deterministic Rules/SQL, classical machine learning, a frozen dense semantic retriever, deterministic relational expansion, and Typed Provenance Graph Retrieval (TPGR), a default-deny typed traversal that admits only persisted transaction relationships. Structured baselines are strong: Rules/SQL reaches 84.97% held-out exact accuracy and classical ML reaches 95.44%. Holding the reasoning model, prompt, and generation configuration fixed and changing only retrieval, macro required-record recall moves from 0.83% to 77.70% and exact 16-class accuracy from 2.05% to 72.44% (paired difference 70.39 percentage points; 95% bootstrap CI 66.06–74.72), while TPGR uses fewer records (19.56 versus 40) and fewer source tokens (approximately 2,370 versus 5,531). Decomposing the outcome shows that structural retrieval failures outnumber reasoning failures with sufficient retrieval by 95 to 15. Yet 254 correct predictions occur despite incomplete retrieval, and strict returned-evidence contract accuracy is only 5.72%. On FinRCA-Bench, retrieval architecture rather than model capability determines the dominant share of end-to-end performance — and a correct root-cause label is a weak proxy for an auditable one.

## 1. Introduction

Retrieval-augmented language systems are increasingly asked to reason not over documents but over the operational records that enterprises actually run on [1]. This shift changes what relevance means. In document settings a passage is relevant because of what it says; in transactional settings a record is often relevant because of a relationship it participates in, and it may share almost no content with the question being asked. The consequence for evaluation is that end-to-end accuracy becomes ambiguous. A system that answers incorrectly may have reasoned badly, or it may simply never have been shown the records that would have made the answer derivable. Separating these two failure modes requires knowing, independently of the answer, which records the task actually needed.

Financial operations are an unusually sharp instance of this problem, and one in which the distinction carries practical weight. Evidence about a single transaction is distributed across invoices, purchase orders, approvals, allocations, payments, ledger entries, bank activity, vendor history, and audit records, connected by persisted transactional relationships rather than by textual similarity. An allocation row can be decisive because it bridges an invoice and a payment; a general-ledger entry because its typed source transaction establishes accounting consequence; a vendor-change event because it altered the entity party to the transaction. Two records can therefore be jointly necessary to an investigation while sharing little text, and two records can look semantically or numerically similar without constituting authoritative financial provenance. Financial settings also raise the evidentiary bar: a finance user acts on the explanation and the records behind it rather than on a predicted label, so a diagnosis that cannot be traced to specific source records has limited operational value.

We study this problem through reconciliation root-cause analysis (RCA). Reconciliation is often described as a matching problem — determining whether two records agree on amount, date, counterparty, or reference — but in operational systems a mismatch is usually only the symptom. The investigation must determine why the mismatch occurred and which records establish that explanation. A payment that appears in an ERP but not in bank activity may still be within a legitimate clearing window, may have failed in transmission, may reflect a vendor-master conflict, or may have been linked to the wrong bank transaction. These causes are distinguished not by how the anchor record reads but by which other records exist, what they contain, how they are connected, and occasionally by what does not exist at all.

Reconciliation is also costly in a way that is specific to evidence. Practitioner accounts of the monthly close describe the labour as dominated not by deciding what went wrong but by assembling the records needed to decide: exporting statements from bank portals, merchant processors, and sub-ledgers, then normalising formats and identifiers before any matching can begin, with exception investigation — the diagnostic step itself — consuming the smallest share of the effort [17]. Our experiments locate the same asymmetry in an automated system: on FinRCA-Bench, structural retrieval failures outnumber reasoning failures with sufficient evidence by 95 to 15. We claim no quantitative correspondence between human effort and model failure. We note only that both concentrate in evidence assembly rather than in diagnosis, which is the asymmetry this paper is built to measure.

This setting differs from much of financial language-model evaluation, in which the evidence boundary is supplied or is semantically addressable. FinQA and TAT-QA focus on numerical reasoning over financial reports and tables [2], [3]; FinanceBench evaluates open-book question answering over public-company filings [4]; FinAgentBench and FinRetrieval examine retrieval over financial documents or structured data [5], [16]; Finch evaluates long-horizon spreadsheet-centric finance workflows [6]; and FinAuditing, FinRule-Bench, and FinBalance study structured auditing, rule diagnosis, and multi-document accounting reconciliation [7]–[9]. These benchmarks substantially advance financial AI evaluation and investigate problems that the present work does not. What they do not jointly instantiate is transaction-level RCA over a multi-table operational lifecycle with evaluator-private record-level evidence contracts and a controlled retrieval intervention — the combination that makes evidence access measurable separately from reasoning.

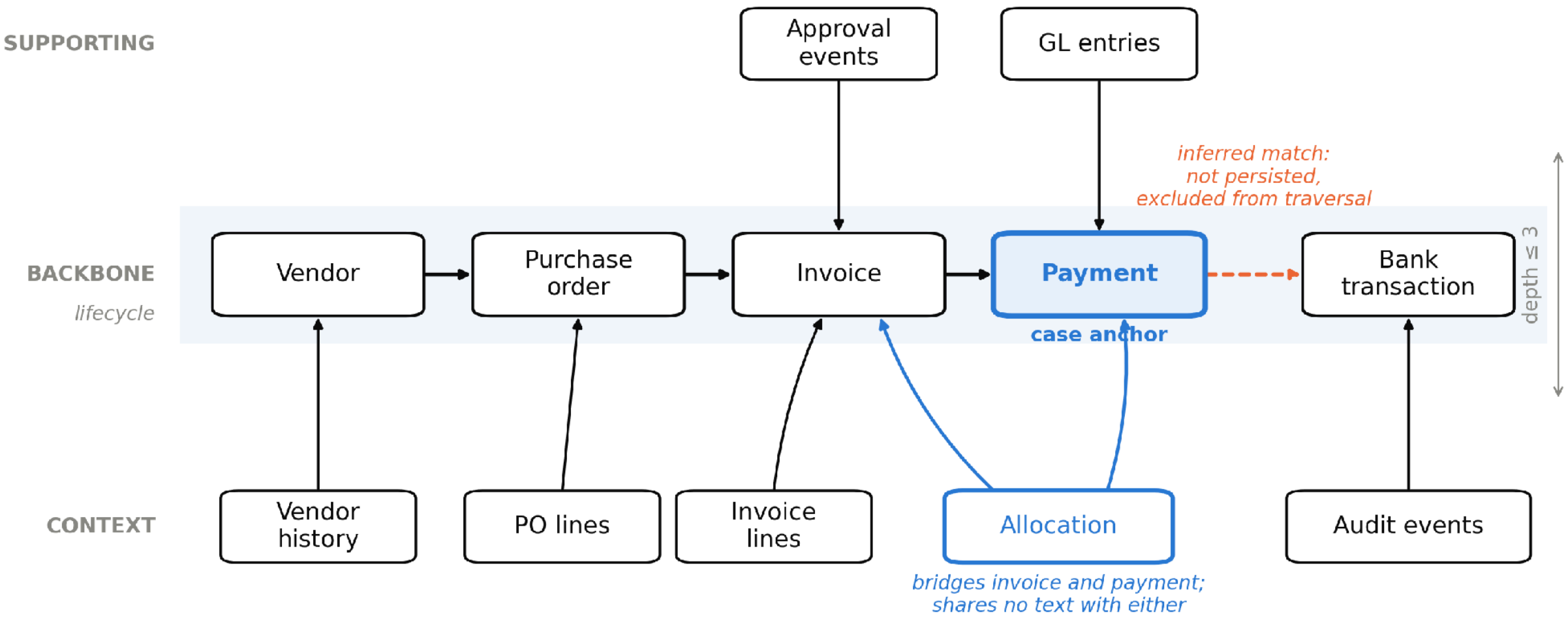

**Figure 1.** Operational financial evidence is organised by persisted transaction relationships rather than by content. Backbone lifecycle records (shaded) are linked to supporting and context records by typed edges, which TPGR traverses to a maximum depth of three (Section 5.5). Two properties drive the study. An allocation row is decisive because it bridges an invoice and a payment while sharing almost no text with either, so it is not reachable by content similarity. The payment-to-bank correspondence is an inferred match rather than a persisted relationship and is therefore excluded from traversal, which is the design decision behind the zero held-out recall on class F15 (Section 7.6).

Graph-based retrieval is a natural candidate in this setting, but our aim is not to claim that graph retrieval is universally superior to retrieval-augmented generation. Existing GraphRAG and financial HybridRAG work primarily builds or uses graphs for question answering over unstructured documents [12], [13]. Enterprise graph-grounded benchmarks such as ENTLORE show that making implicit organizational relations navigable can improve enterprise QA [10], while OpenRCA 2.0 demonstrates in a non-financial domain that outcome-only RCA evaluation can conceal failures in the causal path [11]. We therefore treat our graph retriever as a controlled instrument rather than as a proposed system: because its traversal is forbidden from using amount, date, currency, or fuzzy-identifier similarity, a difference in what it recovers cannot be attributed to a better similarity function. This narrows the question to one we can answer: what happens when a financial RCA system must recover explicit operational provenance before the reasoning model observes the case?

**Novelty boundary.** Among the financial, enterprise, and RCA benchmarks reviewed for this manuscript, we are not aware of prior work that jointly combines (i) transaction-level causal RCA over a multi-table financial lifecycle, (ii) evaluator-private record-level evidence contracts derived from known injected causes, (iii) an explicit distinction between persisted provenance and hypothesized links, and (iv) a controlled held-out retrieval intervention that holds the downstream LLM configuration fixed. FinRCA-Bench is designed to study this intersection rather than to replace existing financial QA, auditing, or agent benchmarks.

We make four contributions.

- **FinRCA-Bench, a benchmark that makes evidence access measurable.** It contains 2,250 deterministic cases over a 14-table, 155,391-row accounts-payable-to-bank corpus, with 15 injected causal failure categories, 750 hard negatives, causal-vendor-group split isolation, and evaluator-private record-level evidence contracts derived from known injected causes. Unlike prepared-context financial QA, the required evidence set is known to the evaluator and hidden from the system, so retrieval can be scored without reference to the answer.
- **An evidence-ladder evaluation and a retrieval-versus-reasoning attribution protocol.** Four levels — outcome correctness, evidence access, evidence sufficiency, and semantic support — of which the first three are measured automatically and the fourth is defined but left unadjudicated. The attribution protocol partitions every case into structural retrieval failure, reasoning failure with sufficient retrieval, success with sufficient retrieval, success despite incomplete retrieval, or technical failure, under the standing commitment that a correct label never retroactively certifies retrieval. The protocol is domain-independent.
- **A controlled held-out demonstration that retrieval architecture, not model capability, dominates end-to-end performance in this setting.** With the reasoning model, prompt, taxonomy, output schema, and generation configuration held fixed and retrieval as the only changed component, macro required-record recall moves from 0.83% to 77.70% and exact 16-class accuracy from 2.05% to 72.44% (paired difference 70.39 percentage points; 95% bootstrap CI 66.06–74.72), with the graph retriever using fewer records and fewer source tokens. Typed Provenance Graph Retrieval (TPGR) serves as the instrument: its default-deny design admits only persisted transaction relationships, which is what licenses reading the difference as an evidence-access effect.
- **Evidence that correct diagnosis and grounded diagnosis are different achievements.** Structural retrieval failures outnumber reasoning failures with sufficient retrieval by 95 to 15; 254 correct predictions occur despite incomplete retrieval; and in 311 of 437 cases (71.17%) a correct class is returned with evidence that fails the contract, with strict returned-evidence contract accuracy at 5.72%. Class-level attribution localizes the residual failures to three identifiable gaps in retrieval operators rather than to reasoning.

The headline result is therefore not that an LLM outperforms conventional finance models. It does not: classical ML reaches 95.44% held-out exact accuracy, compared with 72.44% for TPGR + LLM. The two are also not doing the same job — the classifier assigns a label from construction-aligned structured features, whereas the retrieval-and-reasoning system must locate its own evidence and return it. The central result is that the same LLM moves from 2.05% accuracy under frozen Dense RAG to 72.44% under TPGR as macro required-record recall moves from 0.83% to 77.70%, and that decomposing the remaining errors attributes 95 of them to structural retrieval failure against 15 to reasoning with sufficient evidence. For source-record investigation, retrieval architecture determines what financial world the reasoning model is permitted to observe.

## 2. Related Work

Table 1 summarizes the closest benchmark families. The comparison is intentionally property-based rather than a priority claim: different benchmarks target different units of work, and several are complementary to FinRCA-Bench. The column that most sharply separates them is the granularity of evidence supervision — whether the evaluator holds a ground-truth set of individual source records against which retrieval can be scored independently of the answer.

**Table 1.** Positioning against representative financial, enterprise, and RCA benchmarks. The comparison is property-based and is not a priority claim; several of these benchmarks are complementary to FinRCA-Bench. The distinguishing property of FinRCA-Bench is evidence supervision at the level of individual source records over an operational transaction lifecycle, not the reconciliation domain itself.

| Benchmark | Primary setting | Core task | Evidence / structure supervision | Relation to FinRCA-Bench |
|---|---|---|---|---|
| FinQA / TAT-QA [2], [3] | Financial reports and tables | Numerical QA | Programs / answer evidence | Prepared artifacts; not operational RCA |

| Benchmark | Primary setting | Core task | Evidence / structure supervision | Relation to FinRCA-Bench |
|---|---|---|---|---|
| FinanceBench [4] | Public-company filings | Open-book QA | Evidence strings | Document QA rather than transaction provenance |
| FinAuditing [7] | XBRL multi-document statements | Semantic, relational, numerical audit checks | Taxonomy-aligned structure | Auditing disclosures; not transaction-lifecycle RCA |
| FinRule-Bench [8] | Financial statements + accounting rules | Rule verification and diagnosis | Rule / violation structure | Diagnostic reasoning, but statement-level rather than operational lifecycle |
| FinBalance [9] | Source-document bundles | Journal construction, balance sheet, inconsistency detection | Support-document links | Accounting reconciliation, but not multi-table operational RCA |
| ENTLORE [10] | Enterprise documents and records | Latent organizational QA | Private truth graph / proof certificates | Enterprise relational reasoning; not financial causal RCA |
| OpenRCA 2.0 [11] | Software telemetry | System root-cause analysis | Stepwise causal propagation path | Closest process-level RCA analogue, non-financial |
| FinRCA-Bench | Operational financial records | 16-class financial RCA | Required records + evidence contracts | Transaction-level causal RCA with controlled retrieval |

## 2.1 Financial reasoning and numerical question answering

FinQA and TAT-QA established financial reasoning over mixed textual and tabular evidence, evaluating whether a model can execute the numerical program that a prepared context supports [2], [3]. FinanceBench extended this to open-book financial QA over public-company filings with reference evidence strings [4]. In each case the evidence boundary is supplied or is semantically addressable: a question about a financial concept is answered by content that discusses that concept. The capability under test is reasoning over evidence rather than reconstruction of the evidence set, and retrieval failure and reasoning failure are consequently difficult to separate.

## 2.2 Financial benchmarks for auditing, reconciliation, and enterprise workflows

More recent work moves closer to back-office finance. FinAuditing evaluates semantic, relational, and mathematical consistency over taxonomy-structured financial disclosures [7]. FinRule-Bench studies diagnostic completeness under explicit accounting principles [8]. FinBalance evaluates source-document reconciliation into journal entries and balance sheets and reports that plausible numerical outputs can still be bound to the wrong supporting documents [9] — an observation closely related to the grounding gap we measure. Finch broadens the unit of evaluation to long-horizon spreadsheet-centric enterprise workflows [6]. These benchmarks operate at the granularity of statements, disclosures, or document bundles. FinRCA-Bench is complementary: its object of study is the causal failure of an operational transaction process after interdependent records have already been generated across systems.

## 2.3 Retrieval-augmented generation and financial retrieval

Retrieval-augmented generation established the pattern of conditioning generation on retrieved evidence [1], and subsequent financial work has treated retrieval as a substantive system component rather than a transparent precursor to reasoning. FinAgentBench frames financial retrieval as a multi-step reasoning problem over document types and passages [5], while FinRetrieval shows that tool and interface availability can dominate agent performance on structured financial lookup [16]. In all of these settings relevance remains a property of content, and retrieval quality is ultimately judged by answer quality. The present work makes relevance a property of persisted transactional relationships and scores retrieval directly against a known required-record set.

## 2.4 Graph and relational retrieval for structured reasoning

GraphRAG organizes evidence through graph structure induced largely from prose [12], and HybridRAG demonstrates that graph and vector retrieval can complement one another for financial document QA [13]. Enterprise graph-grounded benchmarks such as ENTLORE evaluate whether latent organizational relations become navigable rather than merely whether documents are retrieved [10]. Our setting differs because the graph is not induced from text: nodes are operational records and edges are frozen transaction relationships that already exist in the source systems. This allows provenance to be treated as a first-class retrieval primitive, and it allows persisted relationships to be held separate from hypothesized entity matches.

### 2.5 Evidence attribution and process-level evaluation

Attributed Question Answering and ALCE separate answer correctness from attribution quality [14], [15]. OpenRCA 2.0 is especially relevant conceptually: it argues that outcome-only RCA can conceal an “ungrounded diagnosis” when a system names the right cause without recovering a verified propagation path [11]. FinRCA-Bench adopts the same general principle but operationalizes it differently — through record-level evidence contracts over a transaction graph, and through an attribution protocol that assigns each case to retrieval, reasoning, or technical failure. The financial operational-record setting adds constraints that are absent from document attribution: exact transaction identity, absence-based evidence, many-to-many mappings, and the distinction between authoritative and inferred links.

## 3. Problem Formulation and the Evidence Ladder

### 3.1 Reconciliation root-cause analysis

Let D = {r1, ..., rn} denote a corpus of operational records. Each record belongs to an entity type such as vendor, purchase order, invoice, approval event, payment, allocation, GL entry, bank transaction, or audit event. For a reconciliation case c, the system observes a routed case anchor and model-visible records and predicts one label y-hat from 16 classes: 15 failure categories plus NO_FAILURE.

$$D = \{r_1, r_2, \ldots, r_n\}, \quad \hat{y}_c \in Y, \quad |Y| = 16$$

An RCA system should do more than classify. It should recover the records required to investigate the discrepancy, distinguish the causal failure from its observable symptom, and return evidence that supports the diagnosis. Some contracts also include meaningful absence: for example, an ERP payment that should have a bank counterpart but does not. Absence cannot be retrieved by similarity to anything and is therefore a structural stress test for any content-based retriever.

### 3.2 Evidence contracts

For each evidence-evaluable case c, let $G_c$ be the set of required source records under the benchmark's record-presence projection and let $R_c$ be the set returned by a retriever. Per-case record recall is $|G_c \cap R_c| / |G_c|$. Full-record coverage requires $G_c \subseteq R_c$. Because $G_c$ is derived from the known injected cause and is never exposed at inference time, retrieval can be scored without consulting the predicted label. The evaluator also retains stricter evidence contracts that incorporate observable conditions and absence-based requirements; these are intentionally separated from simple record-set coverage.

$$\mathrm{Recall}(c) = |G_c \cap R_c| / |G_c|, \quad \mathrm{FullRecord}(c) = 1[G_c \subseteq R_c]$$

### 3.3 The evidence ladder

We operationalize evidence-grounded RCA as a four-level ladder. Level 1 is outcome correctness: was the RCA class correct? Level 2 is evidence access: did retrieval expose the required operational records? Level 3 is evidence sufficiency: did the returned evidence satisfy the benchmark contract? Level 4 is semantic support: do the cited records actually support the model's causal statements? The levels are scored independently and are not interchangeable; in particular, a system may pass Level 1 while failing Levels 2 and 3, which is the central empirical observation of this paper. The frozen experiment measures Levels 1–3 automatically; Level 4 remains unadjudicated and is therefore not assigned a numerical score. Figure 2 shows the held-out TPGR result at each level.

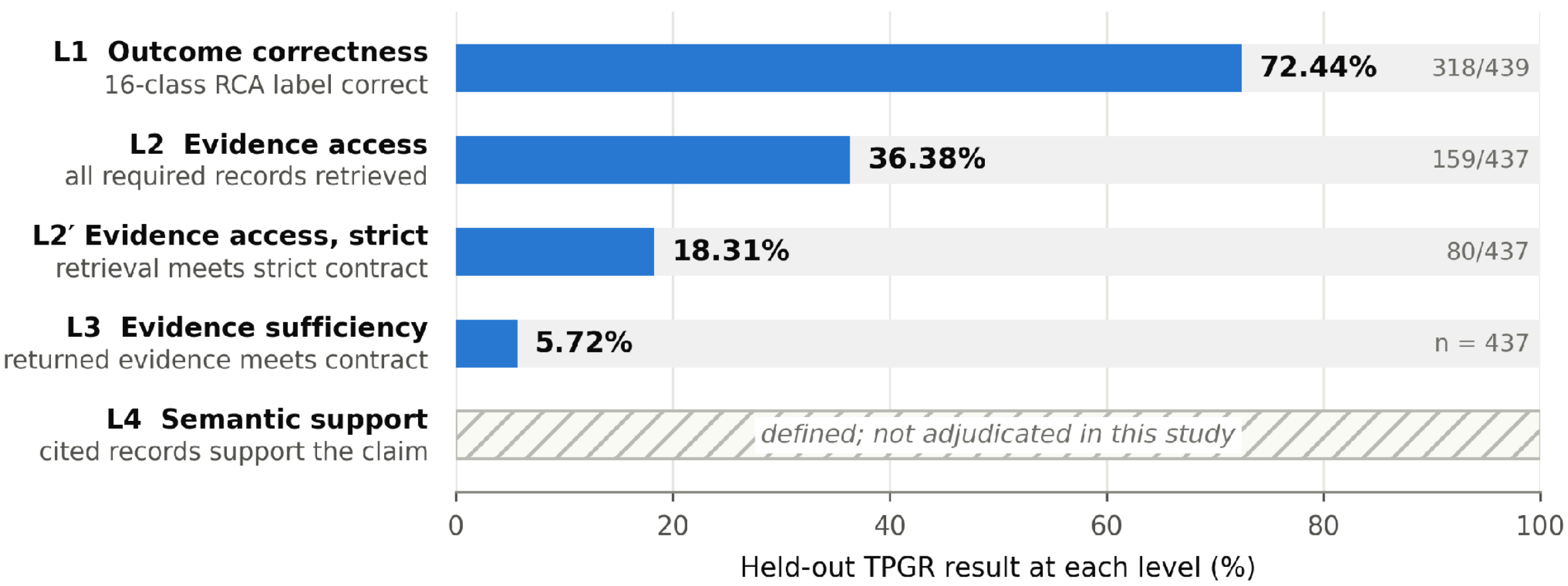

**Figure 2.** The evidence ladder scores four properties of an RCA decision independently. Bars show held-out TPGR results at each level. The levels use different denominators and measure different constructs, so the bars are not comparable to one another as metrics; the quantity of interest is the monotone collapse from 72.44% outcome correctness to 5.72% strict returned-evidence sufficiency. Level 4 is defined by the framework but was not adjudicated in this study and is shown unfilled.

### 3.4 Research questions

- **RQ1:** How accurately do structured deterministic and supervised methods diagnose the frozen FinRCA-Bench failure taxonomy when given construction-aligned access to the operational schema?
- **RQ2:** When required evidence is defined by transactional relationships rather than content, how much of it does a frozen dense semantic retriever recover?
- **RQ3:** Does typed multi-hop traversal over persisted relationships recover required evidence that deterministic single-hop relational expansion cannot reach?
- **RQ4:** With the downstream reasoning model held fixed, does improved evidence access translate into better RCA, and does the resulting failure distribution attribute errors to retrieval or to reasoning?
- **RQ5:** Does a correct root-cause label imply that the system accessed and returned sufficient evidence to justify it?

## 4. FinRCA-Bench

### 4.1 Design objective and operational schema

FinRCA-Bench v1.0.0 is a deterministic synthetic benchmark designed to model a selected accounts-payable-to-bank lifecycle. It provides known causal interventions, hard negatives, record-level evidence annotations, and a fully model-visible operational corpus while keeping labels, mutation provenance, and oracle evidence private during inference. Determinism is a design requirement rather than a convenience: knowing the injected cause exactly is what permits an enumerable required-evidence set, and an enumerable required-evidence set is what permits retrieval to be scored apart from reasoning. All organizations, people, identifiers, transactions, and monetary values are synthetic; no private enterprise or bank data are used.

**Table 2.** FinRCA-Bench v1.0.0 summary. Counts are post-injection; the complete per-table breakdown is given in Appendix A.

| Property | Value |
|---|---|
| Cases | 2,250 total: 1,500 failures + 750 NO_FAILURE / hard-negative cases |
| Failure taxonomy | 15 causal categories, 100 injected cases per category |
| Operational tables | 14 |
| Core row counts | 500 vendors; 5,000 POs; 8,147 invoices; 6,200 payments; 7,180 allocations; 46,765 GL entries; 6,263 bank transactions |
| Splits | 1,192 train; 416 validation; 439 held-out test; 203 challenge test |
| Dataset SHA-256 | c73ad3e9…92761d8 (full hash reported in Reproducibility statement) |

The 14 operational tables are vendors, vendor_change_log, purchase_orders, po_lines, invoices, invoice_lines, approval_events, payments, payment_allocations, gl_entries, bank_transactions, bank_statements, employees, and audit_log. Their relationships include PO-to-line, PO-to-invoice, invoice-to-line, invoice-to-approval, invoice-to-allocation, payment-to-allocation, typed GL source transactions, vendor history, bank-statement membership, and employee or audit references. These relationships are persisted in the source systems; they are not inferred, and the benchmark treats that distinction as load-bearing.

### 4.2 Deterministic construction and causal injection

Generation proceeds in two stages. A clean financial world is first created in lifecycle order and must pass structural, temporal, accounting, and bank-statement checks. Deterministic failure injectors then mutate eligible entity groups, propagate consequences where appropriate, record the causal mechanism, and generate evaluator-private RCA ground truth. Failure injectors claim their primary entity groups to prevent later injected failures from reusing the same causal group.

The generator preserves accounting invariants. Monetary calculations use fixed-point decimal arithmetic; journals remain double-entry balanced even when a failure is economically wrong; and bank statements are recomputed after bank-side mutations so that opening balance + credits − debits = closing balance. The aim is to avoid trivial corruption artifacts that would let a model identify a failure without reconstructing its financial cause.

### 4.3 Failure taxonomy and hard negatives

**Table 3.** Frozen causal failure taxonomy. Each category receives exactly 100 injected cases (1,500 in total); the remaining 750 cases are NO_FAILURE hard negatives, giving a 16-class prediction target.

| ID | Failure category | Primary lifecycle area |
|---|---|---|
| F01 | Duplicate invoice | Invoice |
| F02 | PO / invoice amount mismatch | Procurement |
| F03 | Quantity mismatch | Procurement |
| F04 | Incorrect vendor association | Vendor / procurement |
| F05 | Payment without valid invoice | Payment |
| F06 | Invoice paid twice | Payment |
| F07 | Partial payment / residual balance | Payment |
| F08 | Approval workflow failure | Workflow |
| F09 | GL posting mismatch | Accounting |
| F10 | Wrong accounting period | Accounting |
| F11 | Vendor master change conflict | Vendor master |
| F12 | ERP payment missing from bank | Cross-system settlement |
| F13 | Bank transaction missing from ERP | Cross-system settlement |
| F14 | Bank / ERP amount mismatch | Cross-system settlement |
| F15 | Incorrect payment-to-bank match | Cross-system matching |

The 750 NO_FAILURE cases include legitimate near-misses such as amount differences inside tolerance, scheduled split payments, multi-invoice batch settlements, payments within valid clearing windows, documented vendor changes, properly journaled bank fees, reversals and cancellations, and legitimate foreign-exchange or allocation differences. These cases are designed to break simple heuristics such as “different amount implies failure” or “multiple payments imply duplicate payment,” and they ensure that a system cannot succeed by treating any observed discrepancy as a failure.

### 4.4 Relational complexity and evidence depth

The benchmark intentionally includes cases for which required evidence is reachable only through multi-hop relationships, so that evidence assembly rather than local matching is the binding constraint. The frozen deterministic rule registry classifies 9 of 15 rules as Tier-3 relational reasoning, with registered minimum paths extending to four hops. In the independent 416-case TPGR validation analysis, 156 cases contained required evidence reachable only at graph depth two or three; 239 required record incidences occurred at depth two and 43 at depth three. Multi-hop traversal converted 88 validation cases into fully covered record sets.

### 4.5 Splits, leakage control, and quality assurance

Cases are split by causal vendor group rather than by individual record. A deterministic seed-dependent hash assigns the entire primary vendor group to train, validation, held-out test, or challenge test. Quality checks report zero pairwise case-ID overlap and zero primary-vendor-group overlap across splits. Model-visible financial data are separated from labels, failure manifests, mutation provenance, and ground-truth-derived evidence packages. The final benchmark quality gate passed all 1,500 injected failure signatures, primary-key duplication checks, and clean/post-injection structural checks.

## 5. Methods

All completed methods operate against the same frozen operational snapshot with an analytical cutoff of June 30, 2026. Ground-truth labels, expected evidence, mutation provenance, and failure manifests are prohibited from inference-time inputs. The methods are not intended as competitors for a single leaderboard: the structured baselines establish what the taxonomy permits, and the retrieval arms isolate what evidence access contributes.

### 5.1 Deterministic Rules/SQL

The deterministic baseline contains 15 rules, one per failure category. Rules use exact joins and anti-joins, aggregation, decimal arithmetic, temporal conditions, normalization, and a narrowly specified invoice-reference comparison. Each rule returns ANOMALY, MATCH, or INSUFFICIENT_EVIDENCE, and a frozen collision policy maps simultaneous rule outcomes to a single RCA label while retaining raw traces. The only preregistered validation search selected the F01 duplicate-invoice window (7 days) and normalized-reference threshold (1.00); all other thresholds were frozen before held-out evaluation. This baseline is given privileged schema access by design: it is a measurement of what the taxonomy permits under construction-aligned logic, not a candidate for deployment on an unfamiliar ledger.

### 5.2 Classical machine learning

The ML baseline represents each case with 147 pre-encoding structured features: 27 categorical, 54 raw or aggregate numeric, 20 difference, 26 match/comparison, 15 missingness, and 5 relational features. Train-fitted preprocessing yields 265 transformed features. Twenty-two preregistered configurations across logistic regression, random forest, and histogram gradient boosting were compared using validation macro F1 under a frozen tie-break hierarchy. The selected estimator was balanced histogram gradient boosting (learning rate 0.1, 200 iterations, 15 leaf nodes, L2 = 1.0, seed 314159). It was not refit on validation data before test evaluation. Like the rule engine, this baseline consumes features engineered around the same taxonomy the benchmark injects, and it produces a class prediction but no source-record evidence by design. It therefore cannot be placed on the evidence ladder.

### 5.3 Dense RAG: frozen semantic baseline

The semantic comparator is referred to here as Dense RAG rather than “standard RAG” to avoid implying that it represents every modern RAG design. It is the frozen Phase-5 semantic baseline. The corpus contains 155,391 operational rows, each serialized as exactly one document. Documents are embedded once with text-embedding-3-small at 1,536 dimensions, normalized, and searched exhaustively by cosine similarity implemented as inner product. Search is exhaustive rather than approximate, so the configuration is not disadvantaged by index error. The case query is deterministic and label-blind. No metadata filtering, lexical retrieval, reranking, entity-aware expansion, known relational neighborhood, or graph hop information is permitted. Validation selects K from {5, 10, 20, 40}; K = 40 is then frozen for held-out evaluation.

### 5.4 Relational Retrieval: deterministic structural comparator

The validation-only structural comparator performs deterministic exact relational expansion from the routed anchor. It can follow permitted operational foreign-key and exact cross-record relationships but does not use TPGR's typed multi-hop traversal grammar or path-first evidence selector. Its purpose is to separate the contribution of relational access in general from the contribution of typed multi-hop traversal specifically. Its output contains 2–21 records per case, under the same maximum-40 evidence-budget contract. No downstream LLM held-out experiment was run for this comparator.

### 5.5 Typed Provenance Graph Retrieval (TPGR) as a controlled instrument

TPGR is the publication name for the frozen Graph v1.1 implementation. It is used here as a controlled instrument rather than proposed as a novel retrieval algorithm: its value in this study is that its restrictions make the resulting comparison interpretable. The underlying graph contains 155,391 source-record nodes and 184,223 edges. The registry contains 22 executable relation types; the frozen traversal grammar permits 30 directed transitions across 15 relation types, grouped as BACKBONE, SUPPORTING, CONTEXT, and TERMINAL_CONTEXT. Traversal depth is capped at three edges and paths cannot repeat record IDs.

The graph uses a default-deny policy: an edge can be traversed only if a frozen grammar entry explicitly permits the current node type, relation, direction, and next node type. Traversal cannot use equal amounts, equal dates, equal currencies, semantic similarity, fuzzy identifiers, unrestricted vendor/employee hub expansion, synthetic journal nodes, or an inferred Payment-to-Bank-Transaction edge. These prohibitions serve two purposes. They preserve the distinction between authoritative operational provenance and a hypothesized match, and they ensure that any recall advantage TPGR displays cannot be attributed to a better similarity function, since TPGR has no similarity function at all.

Candidate paths are compressed by a deterministic, label-blind, path-first selector capped at 40 unique source records. Anchors are mandatory; candidate paths are ordered by a frozen priority lattice favoring backbone accounting and lifecycle paths before supporting or terminal context. Paths are admitted atomically if their marginal set of new records fits the remaining budget. The selector uses no embedding scores, record text, labels, or gold evidence.

### 5.6 Downstream LLM reasoning

Dense RAG and TPGR use the same downstream model configuration: gpt-5.6-sol, medium reasoning effort, the same RCA taxonomy, reasoning prompt, structured output schema, 1,000-token output limit, low verbosity, and retry policy. Retrieval is therefore the only component that varies between the two held-out arms, which is what permits the difference between them to be read as an evidence-access effect rather than a modelling effect. A separately preregistered Direct-LLM arm did not complete held-out execution and is therefore documented in Appendix B and excluded from performance comparisons; as a result, this study does not establish that retrieval is necessary, only that retrieval architecture matters greatly among the arms that completed.

## 6. Experimental Protocol and Metrics

### 6.1 Freeze discipline and evaluation populations

The benchmark, Rules/SQL specification, ML feature registry and model search, Dense RAG protocol and K-selection policy, graph relation registry, TPGR traversal grammar, evidence selector, and downstream reasoning prompt were frozen before their corresponding held-out evaluations. The Graph-vs-Relational experiment uses all 416 validation cases and is retrieval-only. The final TPGR-vs-Dense-RAG LLM comparison uses the canonical 439-case held-out set. Two GL-journal routes cannot be deterministically resolved under the frozen anchor policy and are retained as technical failures for both retrieval systems; retrieval metrics therefore use 437 cases where appropriate while classification accuracy retains all 439 cases.

### 6.2 Classification metrics

The primary end-to-end metric is exact 16-class accuracy. We also report macro F1, binary failure-detection F1, and counts of INSUFFICIENT_EVIDENCE and technical failures. Technical failures remain in the canonical classification denominator rather than being discarded, so that infrastructure error is never silently converted into model error. Paired exact accuracy is compared with an exact two-sided McNemar test; uncertainty in accuracy differences is estimated by paired nonparametric bootstrap.

$$\text{Accuracy} = (1/N)\, \Sigma_i\, 1[\hat{y}_i = y_i]$$

### 6.3 Retrieval and evidence metrics

For evidence-evaluable cases, macro record recall averages per-case required-record recall; micro recall aggregates required record incidences. Full-record coverage requires all required records to be present. Strict full-contract coverage additionally incorporates the benchmark's stricter evidence semantics, including absence-based conditions where applicable. For model outputs, citation-ID validity checks whether a cited ID exists in the provided context; it establishes that the model did not fabricate an identifier, and it does not establish semantic citation support. We keep these two quantities lexically distinct throughout.

### 6.4 Retrieval-versus-reasoning attribution

For TPGR model-context cases, the evaluator independently considers evidence sufficiency and label correctness. Cases are partitioned into structural retrieval failure, reasoning failure given sufficient retrieval, success with sufficient retrieval, success despite incomplete retrieval, or technical failure. We state the governing commitment explicitly, because it is what makes the partition informative: a correct class never retroactively marks retrieval as successful, and an incorrect class is never counted as a reasoning failure unless retrieval was independently judged sufficient. Without this rule, end-to-end accuracy would silently absorb evidence-access failures.

## 7. Results

### 7.1 RQ1: Structured baselines establish that the taxonomy is tractable

Rules/SQL correctly classifies 373 of 439 held-out cases (84.97%). Classical ML correctly classifies 419 of 439 (95.44%), a 10.48 percentage-point improvement. In the paired comparison, ML is uniquely correct on 58 cases and Rules/SQL on 12; exact McNemar $p = 2.25 \times 10^{-8}$. A 2,000-resample paired bootstrap gives a 95% CI of 6.83–14.12 percentage points for the accuracy difference. These results establish that the benchmark is learnable from structured financial information, and therefore that poor retrieval-based performance is not simply a consequence of an impossible taxonomy.

These figures should be read as an upper reference rather than as a production estimate, and we state the reason rather than leaving it to inference. Both structured baselines are aligned with the generative process: the rule engine contains exactly one rule per injected failure category, and the ML feature registry was engineered around the same taxonomy. Neither system must discover which records are relevant, because both are handed the schema. Their strength is informative precisely because it removes one explanation for the retrieval results that follow — the task is decidable once the right records are in hand, so the interesting question becomes whether a system can get them there.

**Table 4.** Structured held-out baselines (n = 439). Higher is better for all three metrics; the best exact accuracy is shown in bold. Both methods receive construction-aligned access to the operational schema and neither returns source-record evidence, so neither can be scored above Level 1 of the evidence ladder. Confidence intervals were not reported for these two arms.

| Method | Split / N | Exact accuracy | Macro F1 | Binary F1 | Source-record evidence |
|---|---|---|---|---|---|
| Rules/SQL | Held-out / 439 | 84.97% (373/439) | 95.18% | 98.01% | Rule traces / evidence |
| Classical ML | Held-out / 439 | **95.44% (419/439)** | 96.24% | 96.63% | None by design |

### 7.2 RQ2: Dense semantic retrieval recovers almost none of the required evidence

Dense RAG returns exactly 40 records per evaluable case but recovers almost none of the benchmark-required evidence. On 437 held-out retrieval cases, macro record recall is 0.83%; micro recall is 29/2,706 = 1.07%; hit rate is 6.41%; and full-record coverage is 0/437. The average retrieved context contains 39.93 irrelevant records out of 40. The downstream LLM returns INSUFFICIENT_EVIDENCE on 428 of 439 cases and reaches only 2.05% exact 16-class accuracy (9/439; 95% CI 0.91–3.42%). All nine exact classifications are NO_FAILURE cases.

Two aspects of this result matter beyond the headline number. First, the failure is localized rather than aggregate: a 6.41% hit rate with zero full coverage says that the retrieved context was not marginally insufficient but categorically misaligned, and the context budget was not the binding constraint — TPGR later reaches 77.70% macro recall using 19.56 records against Dense RAG's 40. Second, the downstream behaviour under misaligned evidence is itself informative: the model abstained on 428 of 439 cases rather than confabulating a cause, and every case it did classify was a negative. Under this configuration the reasoning model was conservative when starved of provenance.

This result should be interpreted as a failure of this frozen dense semantic baseline, not as a claim that all semantic or hybrid RAG systems fail on financial data. We can also be precise about where a stronger baseline would likely help and where it would not. A lexical or hybrid retriever would plausibly recover one-hop identifier co-occurrences, such as an invoice reference appearing verbatim in an allocation row, and would be expected to outperform this baseline on that subset. The harder residual is the depth-two and depth-three requirement — 239 and 43 required-record incidences respectively in the validation analysis — involving records that share no identifier with the case anchor. The general observation is that a large context budget does not help when its contents are misaligned with transaction provenance.

### 7.3 RQ3: Typed multi-hop traversal recovers evidence relational expansion cannot reach

On the 416-case validation retrieval comparison, Relational Retrieval already achieves 68.88% macro record recall, confirming that much of the gap to Dense RAG is attributable to relational access in general rather than to graph structure specifically. TPGR increases recall to 78.70%, an absolute improvement of 9.82 percentage points (10,000-resample paired-bootstrap 95% CI 8.36–11.30). Full-record coverage rises from 63/416 (15.14%) to 151/416 (36.30%). TPGR is better than Relational Retrieval on 156 cases, equal on 257, and worse on 3. For full-record coverage, 88 cases are TPGR-only full, none are Relational-only full, and the exact paired p-value is $6.46 \times 10^{-27}$.

**Table 5.** Validation-only retrieval comparison (n = 416). Higher is better for both metrics; the stronger value in each row is shown in bold. Differences are computed from unrounded values and may differ in the final digit from the difference of the rounded values shown. This comparison is retrieval-only: no downstream reasoning arm was run for Relational Retrieval.

| Metric | TPGR | Relational Retrieval | Difference |
|---|---|---|---|
| Macro record recall | **78.70%** | 68.88% | +9.82 pp |
| Full-record coverage | **36.30% (151/416)** | 15.14% (63/416) | +21.15 pp |
| Maximum record budget | 40 | 40 | — |

Multi-hop traversal contributes materially: 156 validation cases contain required evidence reachable only at depth two or three, and TPGR makes 88 cases fully covered through multi-hop retrieval. Candidate-to-selection loss is small: macro recall is 78.91% before selection and 78.70% after selection, with positive loss in only four cases. Residual failure therefore arises mainly from graph reachability and relation design rather than from the 40-record compression step. We state the scope of this finding at the point of claim: the comparison is validation-only and retrieval-only, so it establishes that typed traversal recovers more evidence, not that the additional evidence changes downstream diagnosis.

### 7.4 RQ4: Evidence access changes downstream reasoning

On the held-out test, TPGR achieves 77.70% macro and 77.01% micro record recall, compared with 0.83% and 1.07% for Dense RAG. TPGR provides complete required records for 159/437 cases (36.38%) and satisfies the strict full evidence contract for 80/437 (18.31%); Dense RAG achieves zero on both measures. TPGR also uses less context: 19.56 records and approximately 2,370 source tokens per case on average, versus 40 records and approximately 5,531 tokens for Dense RAG.

With the reasoning model, prompt, taxonomy, schema, and generation configuration held fixed, TPGR + LLM correctly classifies 318/439 cases (72.44%; 95% CI 68.34–76.54%), while Dense RAG + LLM classifies 9/439 (2.05%). The paired difference is 70.39 percentage points (95% paired-bootstrap CI 66.06–74.72). The paired contingency contains 309 TPGR-only correct cases, 9 both-correct cases, 0 Dense-RAG-only correct cases, and 121 both-wrong cases; exact McNemar $p = 1.92 \times 10^{-93}$. INSUFFICIENT_EVIDENCE falls from 428 cases to 46. Because retrieval is the only component that differs between the two arms, this difference is an evidence-access effect. It is not, however, a measurement of graph structure in isolation: TPGR differs from Dense RAG both in using relational structure and in reaching relevant records at all, and the validation comparison in Section 7.3 is the only arm that begins to separate those two factors. Figure 3 summarizes the intervention.

**Table 6.** Controlled held-out comparison in which retrieval is the only changed component; the reasoning model, prompt, taxonomy, output schema, and generation configuration are identical across arms. Classification metrics use n = 439 and retrieval metrics n = 437. Higher is better for the seven quality metrics, and the stronger value is shown in bold; context size, abstention counts, and technical failures are reported as behavioural and cost quantities and are not scored as better or worse.

| Metric | TPGR + LLM | Dense RAG + LLM |
|---|---|---|
| Exact accuracy | **72.44% (318/439)** | 2.05% (9/439) |
| Macro F1 | **71.38%** | 0.74% |
| Binary F1 | **83.46%** | 0% |
| Macro record recall | **77.70%** | 0.83% |
| Micro record recall | **77.01%** | 1.07% |
| Full-record coverage | **36.38% (159/437)** | 0% |
| Strict full-contract coverage | **18.31% (80/437)** | 0% |
| Mean records / context | 19.56 | 40.00 |
| Mean source tokens / context | 2,370 | 5,531 |
| INSUFFICIENT_EVIDENCE | 46/439 | 428/439 |
| Classification technical failures | 11/439 | 2/439 |

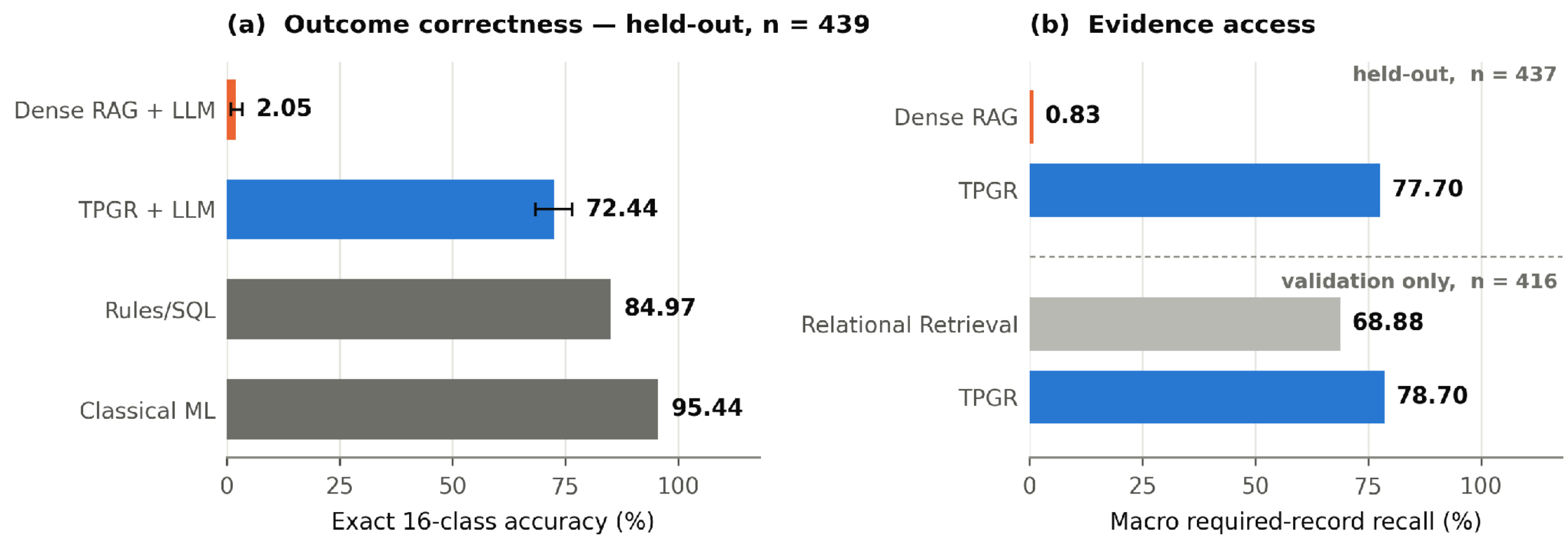

**Figure 3.** The controlled retrieval intervention. (a) Exact 16-class accuracy on the held-out split; the reasoning model, prompt, taxonomy, output schema, and generation configuration are identical for the two LLM arms, so retrieval is the only component that differs between them. Rules/SQL and classical ML (grey) are construction-aligned structured baselines that return no source-record evidence and are shown for reference; classical ML remains the most accurate classifier in the study. (b) Macro required-record recall, held-out above the divider and validation-only below it; the validation pair has no downstream reasoning arm. All values are reproduced from Tables 4-6 without modification.

TPGR + LLM should not be described as the strongest classifier in the study. Classical ML remains substantially better at 95.44% exact accuracy. The contribution of TPGR is different in kind: it operates over explicit source records, provides record-level attributions, and makes it possible to measure whether failures arise from evidence access or from reasoning. A system that cannot expose its evidence cannot be placed on the ladder at all. Of the 11 technical failures recorded in the classification denominator, 9 occur after anchor resolution and appear in the attribution of Section 7.5; the remaining 2 are the unresolvable GL-journal routes described in Section 6.1.

## 7.5 RQ5: Correct labels do not certify grounded diagnosis

The 437 TPGR model-context cases partition into 95 structural retrieval failures, 15 reasoning failures given sufficient retrieval, 64 successes with sufficient retrieval, 254 successes despite incomplete retrieval, and 9 model-context technical failures. Structural retrieval failures therefore outnumber reasoning failures with sufficient retrieval by more than six to one. Read as an error budget, this says that in the arm where the reasoning model performs best, most of what remains to be fixed is not in the reasoning model. Figure 4(a) shows the resulting partition.

**Table 7.** Retrieval-versus-reasoning attribution over the 437 TPGR model-context cases; rows are mutually exclusive and sum to 437. A correct class is never counted as evidence that retrieval succeeded (Section 6.4). The two GL-journal routes that cannot be resolved under the frozen anchor policy are excluded here and are retained as technical failures in the 439-case classification denominator, which is why Table 6 reports 11 technical failures where this table reports 9.

| Attribution | Cases | Interpretation |
|---|---|---|
| Structural retrieval failure | 95 | Required structure was not sufficiently exposed |
| Reasoning failure with sufficient retrieval | 15 | Evidence available; final RCA incorrect |
| Success with sufficient retrieval | 64 | Correct label with sufficient retrieved evidence |
| Success despite incomplete retrieval | 254 | Correct label without complete retrieved evidence |
| Technical failure | 9 | Model-context/API failure after anchor resolution |

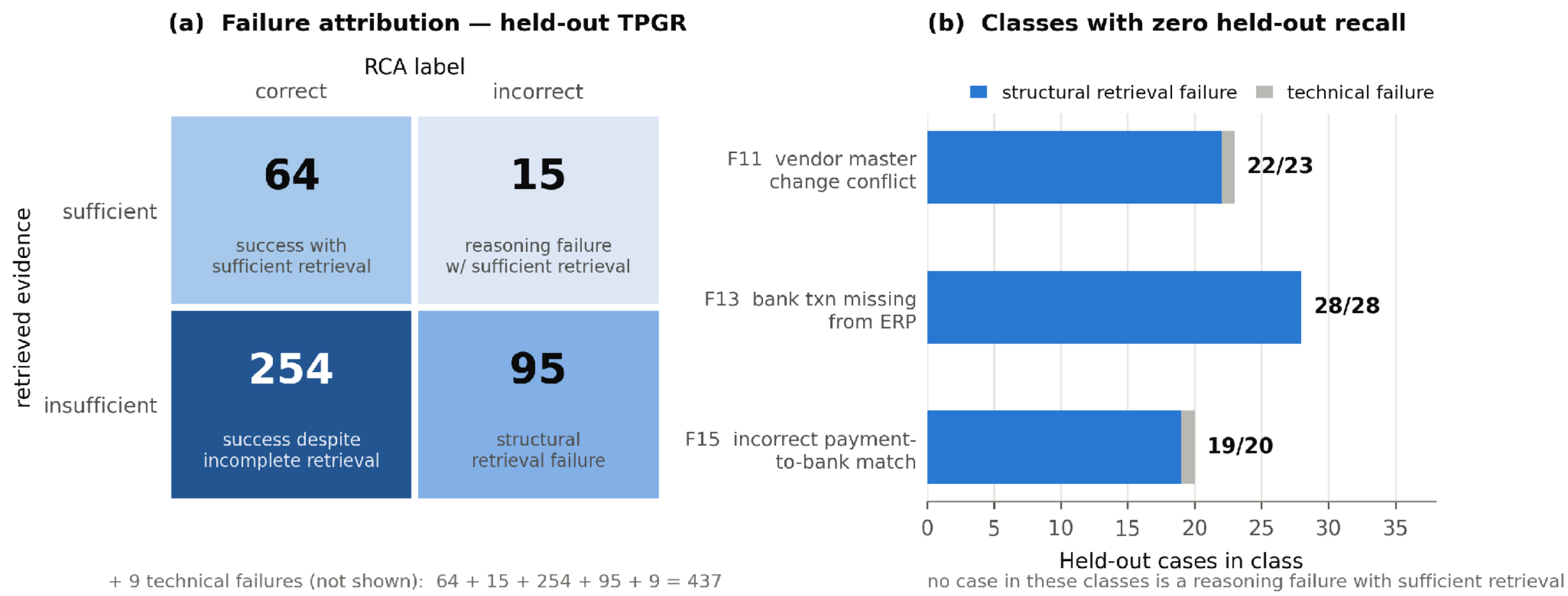

**Figure 4.** Where the failures originate. (a) Every held-out TPGR model-context case partitioned by whether retrieval was sufficient and whether the RCA label was correct; a correct label never retroactively marks retrieval successful (Section 6.4). Structural retrieval failures outnumber reasoning failures with sufficient retrieval by 95 to 15, while 254 correct labels occur without sufficient retrieval. (b) The three classes with zero held-out recall, decomposed by failure type: none of their failures is a reasoning failure with sufficient retrieval, so the residual limitation is in the relation design rather than in the reasoning model. Values are reproduced from Table 7 and Section 7.6.

The model-output evidence evaluation makes the gap sharper. Citation-ID validity is 100% and the hallucinated-record-ID rate is 0%, meaning cited IDs refer to records actually supplied to the model; this is a fabrication check, not a support check. Yet annotated-ID completeness is 21.28%, observable evidence-contract accuracy is 11.44%, and strict evidence-contract accuracy is 5.72%. In 311/437 cases (71.17%), the class is correct while the returned evidence is incorrect or incomplete under the retained evaluator. Semantic citation support and fully grounded reconciliation were not manually adjudicated and are therefore not reported.

The key empirical implication is that label accuracy, retrieval completeness, and evidence sufficiency are not interchangeable measurements. A system reported at 72.44% under Level 1 of the ladder is at 36.38% under Level 2 and 5.72% under the strictest form of Level 3. Any of the three could be quoted as “performance,” and they describe materially different systems.

## 7.6 Class-level attribution localizes the missing retrieval operators

TPGR + LLM has zero held-out recall on F11 (vendor master change conflict), F13 (bank transaction missing from ERP), and F15 (incorrect payment-to-bank match). The failure attribution is highly diagnostic: 22/23 F11 cases, all 28 F13 cases, and 19/20 F15 cases are structural retrieval failures; the remaining cases are technical failures rather than ordinary reasoning failures with sufficient retrieval. F11 exposes the cost of conservative vendor-hub restrictions. F13 exposes the difficulty of representing meaningful absence as a positive graph edge. F15 exposes the deliberate prohibition on manufacturing a Payment-to-Bank edge from amount/date/reference similarity. These are boundaries of the frozen relation design rather than evidence that the reasoning model failed on those classes, and each names a specific retrieval operator that the design lacks. Figure 4(b) decomposes these three classes.

# 8. Discussion

## 8.1 Retrieval is part of the reasoning system

The controlled Dense-RAG-versus-TPGR experiment demonstrates, on this benchmark, that the retrieval layer can dominate observed LLM performance. The same model moves from near-total abstention to 72.44% exact RCA accuracy when its evidence source changes and nothing else does. In relational enterprise settings, retrieval is therefore not a neutral preprocessing stage: it defines which operational facts and relationships are available for downstream reasoning. The corollary is a claim about evaluation rather than about systems, and it is the one we would most like carried forward: evaluations that report only end-to-end accuracy cannot distinguish an evidence-access failure from a reasoning failure, and in this study they would have misattributed 95 failures to the reasoning model.

## 8.2 Why similarity and provenance diverge

Dense semantic retrieval optimizes similarity between a query representation and independently serialized rows. Financial RCA optimizes a different notion of relevance: whether a record is part of the causal transaction path. An allocation row can be essential because it bridges an invoice and a payment, not because its text resembles either. A GL row can be essential because its typed source transaction establishes accounting consequence. A vendor-history event can be relevant because it changed the entity involved in the routed transaction. The held-out result — 77.70% versus 0.83% macro recall despite TPGR using less than half as many records — shows how large this objective mismatch can become in the frozen benchmark. We demonstrate the divergence; we do not claim to have measured how far it extends beyond this corpus.

## 8.3 What the structured baselines do and do not tell us

Classical ML is the strongest classifier in the study. This is not an inconvenient result to be explained away; it is a design lesson. If the sole task is to assign a known RCA label from stable structured features, a non-generative model may be simpler, cheaper, and more accurate. Two qualifications keep the lesson accurate. First, both structured baselines are construction-aligned and schema-privileged, so 95.44% is an upper reference rather than a transferable capability. Second, neither baseline returns source-record evidence, so neither can be evaluated above Level 1 of the ladder. The role our experiments suggest for TPGR + LLM is investigative rather than classificatory: reconstruct source-record context, navigate evidence, and support an analyst-facing explanation. A plausible production design is therefore hybrid — structured controls or classifiers for detection and prioritization, provenance-preserving retrieval for investigation, and an LLM interface for synthesis. That architecture is a hypothesis suggested by these results; we do not evaluate it here.

## 8.4 Correct labels are not sufficient for audit-sensitive RCA

The 254 “success despite incomplete retrieval” cases are central to the paper, and they cut in both directions. They weaken any claim that complete evidence access is necessary for a correct label, and they strengthen the claim that the two quantities are dissociable and must be measured separately. They may reflect redundant diagnostic signals, evidence contracts that are stricter than the minimum needed for statistical classification, or model shortcuts from partial patterns. The current experiment cannot distinguish these explanations, and a minimal-sufficient-evidence ablation would be required to do so. What the result does show is that exact label accuracy is an insufficient proxy for an auditable investigation. A finance user may act on the explanation rather than the label, and that requires knowing whether the decisive records were accessed and whether the cited records actually support the causal statement.

## 8.5 Provenance, absence, and inferred links require different operators

The zero-recall classes identify three design requirements for future financial retrieval systems, each demonstrated as a gap and each proposed as a hypothesis. First, high-degree entity history should be accessible through scoped, time-bounded provenance operators rather than unrestricted hub expansion. Second, absence is first-class evidence and requires anti-joins or expected-counterpart checks rather than positive graph traversal alone; no similarity function can retrieve a record that does not exist. Third, persisted relationships and hypothesized entity matches should remain distinct. A candidate Payment-to-Bank link inferred from amount, date, or reference similarity may be useful, but its confidence and derivation should remain visible rather than becoming indistinguishable from authoritative provenance.

### 8.6 Implications for benchmark design

The evidence ladder suggests that financial and enterprise AI benchmarks should report at least three separable quantities: outcome correctness, evidence access, and evidence sufficiency. When feasible, semantic citation support should be adjudicated as a fourth layer, which this study defines but does not complete. This recommendation is consistent with attribution research [14], [15] and with process-level RCA evaluation [11], but the operational record setting adds constraints that are especially salient in finance: exact transaction identity, absence-based evidence, many-to-many mappings, and the difference between authoritative and inferred links. The dissociation we measure — 72.44% at Level 1 against 5.72% under the strictest Level 3 — is the empirical case for reporting all three.

### 8.7 Structurally analogous enterprise settings

The property that makes this environment difficult is not that it is financial. It is that relevance is defined by persisted relationships between records, that absence carries evidentiary weight, and that a decision must be justified by specific source records. Payment operations, accounting close, financial controls testing, internal and external audit, fraud investigation, compliance review, insurance claims processing, supply-chain reconciliation, and enterprise knowledge systems over transactional records share that structure to varying degrees. We evaluate one lifecycle in one synthetic corpus and make no empirical claim about any of these settings; we describe them as candidates in which the retrieval-versus-reasoning decomposition may be worth applying, and we distinguish that suggestion from everything demonstrated above.

## 9. Limitations and Threats to Validity

### 9.1 Synthetic construction and construction-aligned baselines

FinRCA-Bench is synthetic and models a selected accounts-payable-to-bank lifecycle. Production environments contain schema drift, inconsistent identifiers, undocumented conventions, delayed ingestion, manual adjustments, additional systems, naturally imbalanced incident prevalence, and simultaneous root causes. A related and distinct concern is that the deterministic rule engine and the ML feature registry are aligned with the generative process — one rule per injected category, and features engineered around the same taxonomy — so their accuracy should not be read as evidence that the task is easy in deployment. The reported absolute accuracies are therefore not production estimates. What determinism buys in exchange is the ability to enumerate a required-evidence set exactly, which is what makes the retrieval-versus-reasoning decomposition possible at all; this is a deliberate trade of external validity for internal validity, made once and disclosed here.

### 9.2 Retrieval baselines are intentionally bounded

Dense RAG is a clean semantic baseline, not a claim about the best possible RAG system. It does not use BM25, metadata filters, query decomposition, reranking, entity linking, or hybrid structured retrieval, and such methods may improve substantially on the frozen result — most plausibly on one-hop identifier co-occurrences, less plausibly on the depth-two and depth-three requirements documented in Section 7.3. A hybrid baseline is the single most important missing comparison in this study. Likewise, TPGR is one conservative graph design rather than an optimal financial ontology, and no ablation isolates its traversal grammar, priority lattice, or selector.

### 9.3 Relation completeness and negative evidence

TPGR begins from a relatively clean operational schema and deliberately omits some useful but non-authoritative cross-system links. Real deployments would require entity resolution or probabilistic linkage, and negative evidence requires dedicated operators. F11, F13, and F15 show that these limitations materially affect class coverage.

### 9.4 Relational structure and record access are not fully separated

The held-out comparison isolates retrieval architecture but not the specific contribution of graph structure. TPGR differs from Dense RAG both in exploiting typed relationships and in reaching relevant records at all, so the 70.39-point difference cannot be attributed to structure alone. The validation Relational-versus-TPGR comparison partially addresses this and shows that single-hop relational expansion already recovers 68.88% macro recall, but it is retrieval-only with no downstream arm. A held-out Relational Retrieval + LLM arm, and a depth-ablated TPGR, would be needed to attribute the downstream gain to multi-hop structure specifically.

### 9.5 The evidence contracts encode one notion of sufficiency

The retrieval-versus-reasoning decomposition inherits whatever notion of sufficiency the evidence contracts encode. The 254 cases in which a correct label coexists with incomplete retrieval are direct evidence that this assumption deserves testing: they are consistent with contracts that are stricter than the minimum evidence a classifier needs, and also with model shortcuts from partial patterns. Until a minimal-sufficient-evidence ablation is run, the counts in the attribution table should be read as relative to the benchmark's contracts rather than as an absolute account of what each case required.

### 9.6 Incomplete experimental arms and split usage

The preregistered Direct-LLM arm did not complete held-out evaluation and is excluded from comparative performance claims; consequently this study does not establish that retrieval is necessary, only that retrieval architecture matters greatly among the completed arms. The TPGR-versus-Relational comparison is validation-only and retrieval-only; the held-out downstream comparison is TPGR versus Dense RAG. The 203-case challenge split remains unevaluated and should be preserved for a preregistered future generalization test.

### 9.7 Grounding is not fully adjudicated

Citation-ID validity is not semantic citation support: it establishes only that a cited identifier exists in the supplied context. No cases received the planned two-reviewer manual support adjudication, so citation-support and fully grounded reconciliation rates remain undefined, and Level 4 of the evidence ladder is defined but unmeasured. The 5.72% strict returned-evidence contract score is therefore not a substitute for human evaluation of whether a model explanation is actually supported and actionable.

### 9.8 Model and infrastructure specificity

The completed retrieval-based reasoning experiments use one downstream LLM configuration and one embedding model. Different models may respond differently to incomplete relational evidence; in particular, the clean abstention behaviour observed under Dense RAG may not replicate with a model less inclined to withhold an answer, which would change the failure profile even if it did not change the retrieval conclusion. The final TPGR run also includes technical failures, which are retained in the canonical denominator. The study does not benchmark production latency, cost, provider reliability, concurrency, or recovery behavior.

## 10. Conclusion

This paper used financial reconciliation root-cause analysis to ask a question about AI systems: when evidence must be reconstructed across relational records, how much of a reasoning system's apparent competence is determined by its retrieval layer, and can the two be measured apart? FinRCA-Bench answers the second half by construction. Its 2,250 synthetic cases across 14 operational tables, 15 causal failure categories, hard negatives, split-isolated entity groups, and evaluator-private evidence contracts make the required evidence set known to the evaluator and hidden from the system, so that outcome correctness, evidence access, and evidence sufficiency become three independent measurements rather than one.

The decomposition is what the experiments establish most durably. Holding the reasoning model, prompt, and generation configuration fixed and changing only retrieval moved macro required-record recall from 0.83% to 77.70% and exact accuracy from 2.05% to 72.44%, while using fewer records and fewer source tokens; structural retrieval failures then outnumbered reasoning failures with sufficient retrieval by 95 to 15. An outcome-only evaluation would have attributed all of those failures to the reasoning model. Validation results further show that typed multi-hop traversal improves retrieval beyond deterministic relational expansion, though that comparison remains retrieval-only.

Three results bound the claim. Structured classification is strong — Rules/SQL reaches 84.97% and classical ML 95.44% held-out exact accuracy — so the taxonomy is tractable and classical ML remains the best classifier in the study. The frozen Dense RAG baseline is severely misaligned with the benchmark's relational evidence requirements, at 0.83% macro record recall and 2.05% LLM accuracy, but this is a statement about one evaluated configuration and not about semantic retrieval in general. And TPGR, despite recovering substantially more evidence, still reaches zero recall on three classes and produces 254 correct predictions without complete retrieved evidence, with strict returned-evidence contract accuracy at 5.72%.

The durable finding is therefore not “graph beats RAG.” It is that, for this form of financial RCA, representation and retrieval are part of the reasoning problem, and that a correct label is a weak proxy for an auditable one. A system expected to support remediation or audit must move beyond outcome accuracy toward provenance-preserving evidence access, explicit negative-evidence operators, uncertainty-aware record linkage, and independently evaluated support.

The long-term target is not a model that predicts a reconciliation label. It is a system whose diagnosis can be reconstructed, challenged, and verified from the underlying financial records — and an evaluation practice that can tell the difference between the two.

## Reproducibility and Data Statement

The benchmark is deterministic and synthetic. The frozen dataset identity is SHA-256 c73ad3e98575cb4093b1b3898f759d69c57a97840b4d39661b983e91692761d8. The experimental repository retains frozen method specifications, manifests, predictions, hashes, paired tests, and evaluation artifacts. The code, benchmark generation pipeline, frozen experimental specifications, evaluation scripts, predictions, manifests, and reproducibility artifacts are available at https://github.com/PratikGhawate/FinRCA-AI-Bench. The manuscript reports only completed held-out comparisons and explicitly marks incomplete or validation-only experimental arms. No private enterprise or bank data are used in benchmark generation.

## Appendix A. Operational Tables and Row Counts

**Table A1.** Final post-injection operational row counts. See note D1 in Appendix D regarding the relationship between these counts and the 155,391-row serialized corpus.

| Table | Rows |
|---|---:|
| vendors | 500 |
| vendor_change_log | 186 |
| purchase_orders | 5,000 |
| po_lines | 11,231 |
| invoices | 8,147 |
| invoice_lines | 18,299 |
| approval_events | 17,429 |
| payments | 6,200 |
| payment_allocations | 7,180 |
| gl_entries | 46,765 |
| bank_transactions | 6,263 |
| bank_statements | 57 |
| employees | 120 |
| audit_log | 28,034 |

## Appendix B. Preregistered Direct-LLM Arm

A Direct-LLM baseline was preregistered to test reasoning over deterministic, label-blind case packets assembled from the 14 operational tables without semantic retrieval or graph traversal. The frozen configuration used gpt-5.6-sol with medium reasoning effort and structured JSON output. The retained validation stability run completed 63 of 96 planned calls and only 21 cases completed all three planned repetitions. No completed held-out run exists. For that reason, this arm is reported for transparency but excluded from performance tables, statistical comparisons, and novelty claims. Its absence is material to interpretation: without it, this study cannot establish that retrieval is necessary, only that retrieval architecture strongly differentiates the arms that did complete.

# Appendix C. Review-Sensitive Claim Boundaries

- TPGR versus Relational Retrieval is a validation-only retrieval comparison; no held-out downstream LLM comparison between those two methods was performed.
- TPGR versus Dense RAG is the completed held-out downstream comparison, with the LLM configuration held fixed. It isolates retrieval architecture, not graph structure in isolation.
- Dense RAG is one frozen semantic baseline; the manuscript does not generalize its failure to every semantic, hybrid, or reranked RAG design.
- Citation-ID validity establishes that IDs refer to supplied context records; semantic citation support and fully grounded reconciliation were not manually adjudicated, and Level 4 of the evidence ladder is defined but unmeasured.
- A correct RCA label does not imply complete retrieval or sufficient returned evidence.
- The structured baselines are construction-aligned and schema-privileged; their accuracy is an upper reference rather than a production or transfer estimate.
- The retrieval-versus-reasoning attribution is relative to the benchmark's evidence contracts, which encode one notion of sufficiency.